\documentclass[a4paper]{article}
\usepackage{Odyssey2020}

\usepackage{amsmath,graphicx,multirow}

\title{Improving Diarization Robustness using Diversification, Randomization \mbox{and the DOVER Algorithm}}

\name{Andreas Stolcke${}^\ast$%
\thanks{${}^\ast$Research done while author was with the Speech and Dialog Research Group at Microsoft.}}

\address{Amazon Alexa Speech\\
Sunnyvale, CA, USA\\
{\small\tt stolcke@amazon.com}}

\begin{document}

\maketitle

\begin{abstract}
Speaker diarization based on bottom-up clustering of speech segments by acoustic similarity is often highly sensitive to the choice of hyperparameters, such as the initial number of clusters and feature weighting.  
Optimizing these hyperparameters is difficult and often not robust across different data sets.
We recently proposed the DOVER algorithm 
for combining multiple diarization hypotheses by voting. 
Here we propose to mitigate the robustness problem 
in diarization by using DOVER to average across different
parameter choices.
We also investigate the combination of diverse outputs
obtained by following different merge choices pseudo-randomly in the course of clustering, thereby mitigating the greediness of
best-first clustering.
We show on two conference meeting data sets drawn from NIST evaluations that the proposed methods indeed yield more robust, and in several cases overall improved, results.
\end{abstract}
%
\paragraph*{Keywords}
Speaker diarization, acoustic clustering, robustness, meeting diarization, ensemble classification, randomization, DOVER.

\section{Introduction}
	\label{sec:intro}

Speaker diarization is the task of segmenting and co-indexing audio recordings by speaker.
The way the task is commonly defined \cite{TranterReynolds:ieee2006},
the goal is not to identify known speakers, but to 
co-index segments that are attributed to the same speaker;
in other words, diarization implies finding speaker boundaries and 
grouping segments that belong to the same speaker, and, as a by-product, determining the number of distinct speakers.
In combination with speech recognition,
diarization enables speaker-attributed speech-to-text transcription \cite{FiscusEtAl:nist2007}.

A common approach to diarization, especially when the number of speakers is not known, is agglomerative clustering.
The audio is split into initial segments, which are then successively merged based on acoustic similarity until a stopping criterion is met \cite{AjmeraWooters:asru2003,BetserEtAl:icslp2004,LamelEtAl:mlmi2006}.
The initial clusters can be of short, equal length to capture mostly single speakers, or come from an initial speaker change detection step.
Often a clustering step is followed by a realignment step to match acoustic frames to the revised clusters.
The stopping criterion is usually based on a version of the Bayes information criterion (BIC), a cost function that combines model fit to the data with model complexity \cite{BIC}.
Another approach is to stop clustering early and make final merging decisions using a speaker verification criterion \cite{LamelEtAl:mlmi2006}.

Despite these variations, all commonly used bottom-up clustering approaches for diarization share a greedy, one-best approach when deciding which preliminary clusters to merge.
Clustering is iterative, and at each step
the best (according to a similarity criterion) two clusters are combined. 
Small changes to the hyperparameters of the algorithm that change the decision of best merge during the course of clustering can have large effects on the outputs, making it hard to estimate these hyperparameters robustly from development data.

A possible solution to the robustness problem would be to explore a variety of hyperparameter settings, and average results over them, in a form of ensemble classification \cite{Rokach:air2010}.
The question then becomes how to carry out an averaging or voting over multiple diarization outputs.
Unlike classification or recognition tasks that predict labels or label sequences {\em over a shared vocabulary}, diarization is not amenable to a simple voting among alternative outputs, or a sequence alignment step followed by voting, as in the ROVER algorithm \cite{Fiscus:97}.

The DOVER (diarization output voting error reduction) algorithm was recently
proposed to deal with precisely this problem, although originally motivated 
by a need to reconcile alternative diarizations obtained from multiply audio channels \cite{StolckeYoshioka:asru2019}.
In this paper, we apply DOVER to the combination of multiple diarization outputs derived from a single audio input, obtained by varying several hyperparameters, or by explicitly randomizing hard decisions made in clustering.
The hope is that such a ``diversification'' strategy yields more robust behavior across different 
test sets, and possibly overall improved accuracy.

In Section~\ref{sec:DOVER} we review the DOVER algorithm.
Section~\ref{sec:method} presents the experiment setup for our investigation,
and Section~\ref{sec:results} presents the results.
Conclusions and open questions are given in Section~\ref{sec:concl}.

\section{The DOVER Algorithm}
    \label{sec:DOVER}

\subsection{Rationale and Outline}

Here we give only a high-level description of the DOVER diarization-voting algorithm; a detailed, formal description can
be found in \cite{StolckeYoshioka:asru2019}.
An implementation and examples are available on Github \cite{DOVER:github}.

In order to allow voting among different alternative diarization hypotheses,
the DOVER algorithm first maps the anonymous speaker
labels from the various diarization outputs%
\footnote{Without loss of generality, we can assume the labels 
used in the different diarization outputs to be disjoint.}
into a common label space.
Following the mapping, majority voting among the labels can take place
for each region of audio.
A ``region'' for this purpose is a maximal segment delimited by any of the original speaker boundaries, from any of the input segmentations.
The combined (or consensus) labeling is then obtained
by stringing the majority labels for all regions together.

The key question is how labels are to be mapped to a common label space.
We do so by using the same criterion as used by the diarization error (DER)
metric itself,
since the goal of the algorithm is to minimize 
the expected mismatch between two diarization label sequences.
Given two diarization outputs using labels $A_1, A_2, \ldots, A_m$ and 
$B_1, B_2, \ldots, B_n$, respectively, an injective mapping
from $\{A_i\}$ to $\{ B_j \}$ is found that minimizes the total time duration
of speaker mismatches, as well as mismatches between speech and nonspeech.%
\footnote{Such an optimal mapping can be found efficiently using a bipartite graph matching algorithm.
In our implementation, we invoke the NIST DER evaluation script \cite{dscore}
as {\tt md-eval.pl -M} to save the mapping to a file.}
Any labels that have no correspondence (e.g., due to differing numbers of speakers) 
are retained.
For more than two diarization outputs, a global mapping is constructed 
incrementally:  after mapping the second output to the labels of the first,
the third output is mapped to the first two.
This is repeated until all diarization outputs are incorporated.
Whenever there is a conflict arising from mapping the $i$th output to each of the prior
$i-1$ outputs, it is resolved in favor of the label pairing sharing the longest 
common duration (overlap in time).

\subsection{Speech/nonspeech Voting}

Speech/nonspeech decisions are aggregated by outputting a speaker label if and only if the 
total vote tally for all speaker labels is at least half the total of all inputs,
i.e., the probability of speech is $\geq 0.5$.
(This issue will not arise in experiments reported here because all diarization outputs shared the same speech/nonspeech segmentation.)

\subsection{An Example}

\begin{figure*}[tb]
    \centering
    \includegraphics[width=1.5\columnwidth]{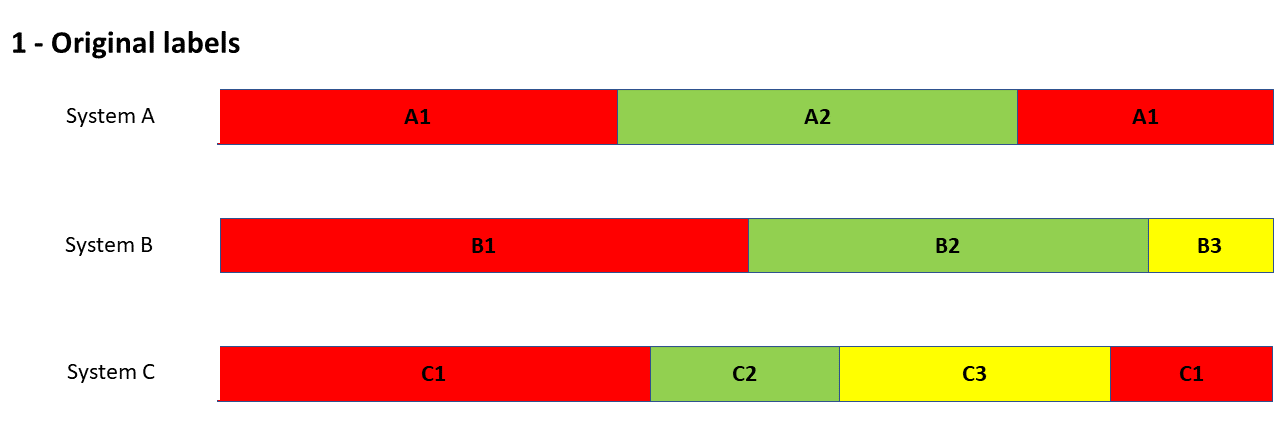}
    \includegraphics[width=1.5\columnwidth]{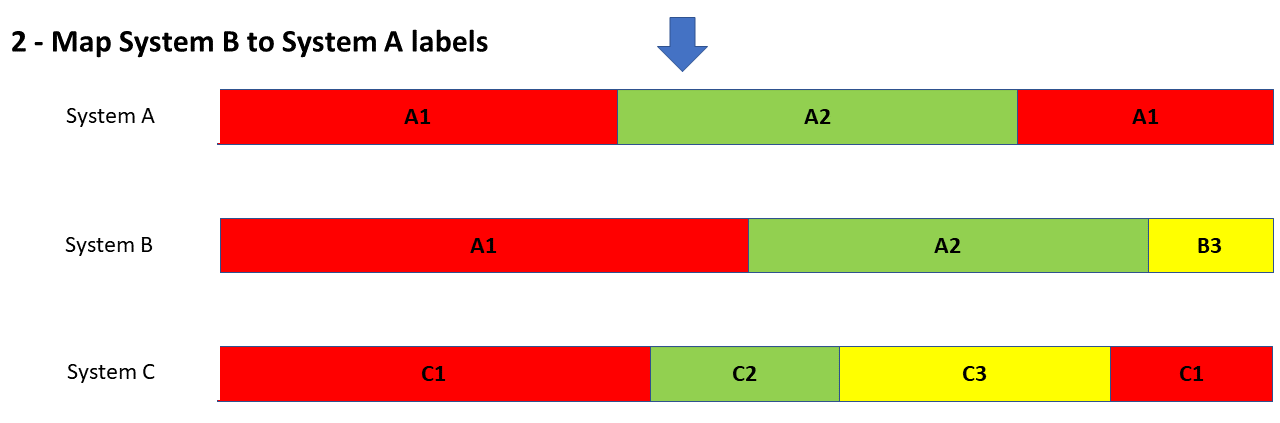}
    \includegraphics[width=1.5\columnwidth]{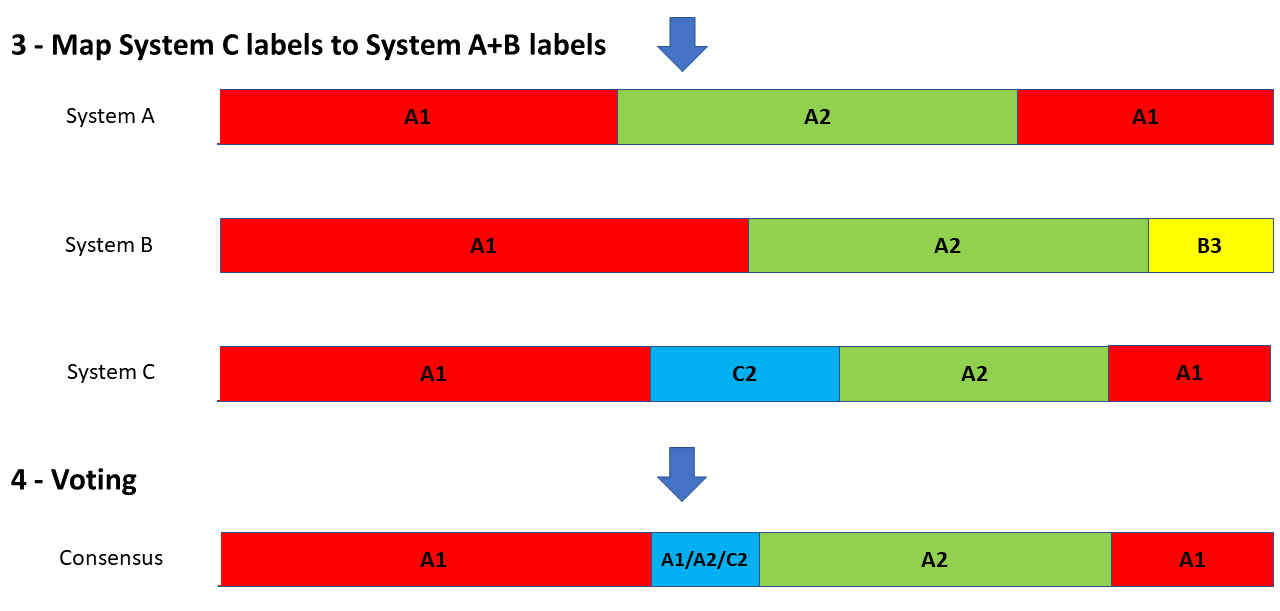}
    \caption{DOVER run on three system outputs (speaker labelings).
            Horizontal extent represents time.
            The original labels are of the form ``B3'', meaning ``speaker 3 from system B''}
    \label{fig:example}
\end{figure*}

An intuitive understanding of DOVER is conveyed by studying an example that is simplified, but also designed to exhibit the algorithm's key features.

Figure~\ref{fig:example} shows how the algorithm processes three diarization hypotheses A, B, and C.
For simplicity, non-speech regions are omitted.
Also for simplicity, the inputs are given equal weight (the generalization to weighted inputs and voting is straightforward).
Step~1 shows the original speaker labelings.
In Step~2 of the algorithm, the labels from System B have been mapped to
labels from System A, using the minimum-diarization-cost criterion.
In Step~3, the output of System C has been mapped to the (already mapped, where applicable) outputs from Systems A and B.

The result of the multiple label mapping steps is that all three diarization versions now use the same labels where possible,
and in the final step (voting) the consensus labels are determined by taking
the majority label for each segmentation region.

Note that the final output contains one region (shown in blue shading) for which 
no majority label exists, since each of the labels ``A1'', ``A2'' and ``C2''
had only one vote.  
A random label could be picked, or the region in question could 
be apportioned equally to the competing labels
(e.g., choosing a temporal ordering that minimizes speaker changes).
Yet another option would be to output a special label denoting uncertain speaker identity or multiple simultaneous speaker labels.
In this paper we break such ties by simply picking the first label.

The algorithm as presented here assumes a single speaker label per input and time interval, and most diarization systems currently do not output labels for overlapping speakers.
However, DOVER could handle overlapping speakers by using sets of speaker labels
(e.g., $\{ \mbox{A1}, \mbox{A2} \}$) as units in voting.
The computation of label mappings requires no changes since the definition of DER already allows for overlapping speakers.

\subsection{Ordering and tie-breaking} 

It may matter in which order the inputs are processed,
and how ties are to be broken in a principled way when no prior weighting of the inputs is given.
As discussed in \cite{StolckeYoshioka:asru2019},
both issues can be addressed by ranking the inputs according to their overall agreement with all the other inputs.  
In experiments described here, we do not employ such a ranking and simply process inputs in a pseudo-random order.

\section{Data and Method}
    \label{sec:method}

\subsection{Data}

\begin{table}[tb]
    \centering
\caption{Statistics of the NIST RT conference meeting sets used}
    \label{tab:rt-stats}
    \begin{tabular}{l|c|c}
    \hline
                    & RT-07 (dev) &   RT-09 (test) \\
    \hline
    No.~meetings    & 8         &   7       \\
    No.~sites       & 4         &   3       \\
    Duration/meeting & 22 minutes  &  19-30 minutes   \\
    Speakers/meeting & 3 - 16  &   4 - 21 \\
    Total speakers  &   31      &   38      \\
    \end{tabular}
\end{table}

We chose two evaluation sets from the NIST Rich Transcription (RT) evaluation series
to validate our approach.
The first dataset was the ``conference meeting'' set from the 
RT-07 evaluation \cite{FiscusEtAl:rt07};
the second set was the corresponding meetings from the RT-09 evaluation.
Both sets had a variety of recording sites, microphone types and numbers, and different numbers of speakers per meeting.
The meetings were hour-long, but only a portion of each meeting was annotated and used for evaluation, for a total evaluation duration of about three hours per set.
Table~\ref{tab:rt-stats} summarizes the statistics for the two data sets.
We used the audio provided for the ``multiple distant microphone'' test condition, 
preprocessed into a single audio stream as described below.

We chose the meeting domain with distant-microphone recordings for several reasons.
First, as is evident from the statistics in Table~\ref{tab:rt-stats},
the datasets are very heterogeneous, especially with regard to the number of speakers. 
Consequently, this data is very challenging for diarization, 
with one manifestation being that it is quite difficult to tune system hyperparameters in a robust way.
Second, we have recently experimented on meeting datasets (and the RT-07 set in particular) when studying multi-channel diarization for meetings, thus providing comparability of results \cite{Denmark:interspeech2019,StolckeYoshioka:asru2019}.

In our experiments we will use RT-07 as the ``development'' or tuning set, and RT-09 as the ``evaluation'' set.

\subsection{Diarization system}

\begin{figure}[tb]
    \centering
    \includegraphics[width=0.75\columnwidth]{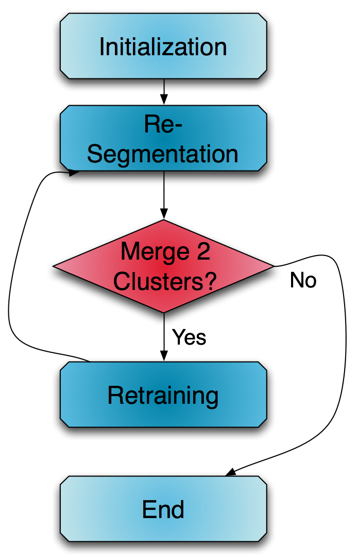}
    \caption{ICSI diarization algorithm flow diagram}
    \label{fig:ICSI-diarization}
\end{figure}

To generate the raw diarization output from audio we used a reimplementation of
the ICSI diarization algorithm \cite{WootersHuijbregts:rt07},
as depicted in Fig.~\ref{fig:ICSI-diarization}.
The algorithm starts with a uniform segmentation of the audio into snippets of equal duration, such that each segment constitutes its own speaker cluster.
This is followed by iterative agglomerative clustering and resegmentation/realignment.
Distance between speaker clusters is measured by the log likelihood difference between
a single-speaker hypothesis (one Gaussian mixture model) versus the two-speaker hypothesis (two GMMs).
In each iteration, the two most similar speaker clusters are merged, 
followed by a resegmentation of the entire audio stream
by Viterbi alignment to an ergodic HMM over all speaker models.
The merging process stops when a BIC-like criterion \cite{AjmeraEtAL:idiap2002} indicates no further gains in the model likelihood.

The acoustic feature front-end was also borrowed from the ICSI RT diarization system,
employing a weighted combination of 19-dimensional Mel cepstra (MFCCs),
and a vector of time delays of arrival (TDOAs) between the different microphones \cite{AngueraEtAl:ieee2007}.
The MFCCs are computed from a single beamformed audio signal, obtained using the BeamformIt tool \cite{BeamformIt}.
The TDOAs are also estimated by BeamformIt, and are combined with the MFCCs by modeling the two feature streams with a weighted combination of separate diagonal-covariance GMMs \cite{AngueraEtAl:ieee2007}.
Because the dimensionality of the TDOA features varies across meetings,
we scale their likelihoods by the inverse of the number of audio channels. 
This way, the relative likelihood contribution of the two streams remains roughly constant, other things being equal.

Speech activity information was obtained from an HMM-based audio segmenter that was part of the 
SRI-ICSI meeting recognition system originally used in the NIST RT-07 evaluation \cite{StolckeEtAl:icassp2010}.
No attempt is made to detect overlapping speech; therefore all our results have an error
rate floor that corresponds to the proportion of overlapped speech.

\subsection{Diversification through Hyperparameter Variation}
    \label{sec:hyperparameters}

We explored two principled ways to generate diverse diarization outputs for the same input.
One method was to vary certain hyperparameters that affect clustering decisions,
specifically
\begin{description}
\item[Stream weight]
    The relative weight given to the TDOA model likelihoods, as compared to the MFCC model likelihoods.
    The value is varied around a point roughly optimized on the development set
\item[Initial clusters]
    The initial number of clusters that are formed by segmenting the input speech at equal-size time intervals.  This value was historically set at 16 in the ICSI algorithm for RT-style meetings.
\item[Number of Gaussians]
    The number of Gaussians allocated to each initial cluster.
    (The number of Gaussians in cluster models increases as clusters are merged.)
\end{description}
We will generally vary these hyperparameter along a grid around default values.

\subsection{Diversification through Randomization}
    \label{sec:randomization}

A second way we can introduce diversity into the diarization result is to explicitly randomize clustering decisions, instead of always merging the two clusters that are
deemed to be closest according to the likelihood criterion.  Specifically, the ICSI algorithm always chooses the pair of clusters for which a merge produces the biggest increase in model likelihood, after pooling and reestimating the Gaussians involved.
(If no merge produces an increase, clustering stops.)
We modify the best-first approach by merging the pair with {\em second} highest likelihood delta with probability $p$, as long as that second best delta is still positive and within relative $L$ of the best.
The latter condition is necessary to rule out very poor merge decisions, which typically lead to bad final results.
Empirically, we found $p = 0.3$ and $L = 1$ give final results that are close (and sometimes better) than those obtained with a strict best-first clustering strategy.
The multiple diarization outputs thus obtained may then be combined with DOVER. 
The spirit of this strategy is similar to random forests of
decision trees \cite{Breiman:2001}, except that diversity is induced by making lower-ranked local choices in a greedy algorithm, rather than selecting subsets of features.

\section{Results}
    \label{sec:results}
    
\subsection{Hyperparameter variation}

In the following experiments, we vary each one of the three hyperparameters described in Section~\ref{sec:hyperparameters}.
In each case, we observe the ensuing variation in results for both dev and evaluation set, as well as the effect of DOVER-voting among the multiple outputs.

Since all experiments use the same speech activity detection output,
results differ only in the speaker error rates, which are reported
in all subsequent tables.
The missed speech rates resulting from the common speech detection algorithm are 3.8\% for RT-07 and 5.8\% for RT-09.
(Note that this includes misses as a result of only hypothesizing a single speaker in regions of overlapping speech.)
False alarm rates are 4.6\% for RT-07 and 4.8\% for RT-09.

\subsubsection{Feature stream weight}

\begin{table}[tb]
    \centering
    \caption{Speaker errors with varying TDOA stream weights}
    \label{tab:results-tdoa}.
    \begin{tabular}{c|c|c}
        \hline
         TDOA weight    & RT-07     & RT-09 \\
                        & (dev)     & (test) \\  
         \hline
            0.715       &   2.9     &   7.7 \\
            0.720       &   \bf 2.6     &   7.5 \\
            0.725       &   2.8     &   9.0 \\
            0.730       &   5.7     &   7.9 \\
            0.735       &   5.7     &   \bf 7.2 \\
            0.740       &   3.7     &   7.9 \\
            0.745       &   \bf 2.6     &   7.7 \\
            0.750       &   4.1     &   8.5 \\
            0.755       &   2.8     &   7.5 \\
            0.760       &   2.8     &   7.5 \\
            \hline  
            DOVER       &   \bf 2.5     &   \bf 7.4 
    \end{tabular}
\end{table}

Table~\ref{tab:results-tdoa} gives results for the TDOA weight hyperparameter. Clustering used 16 initial clusters with 5 Gaussians each.

It jumps out that the results vary in a non-smooth fashion while sweeping the parameter value, and that development set minima (2.6\% error) do not predict optimal results on the eval set (achieved at 0.735, with 7.2\% error).
The DOVER combination yields a slight improvement over the best result on RT-07 (2.6\% $\rightarrow$ 2.5\%), and is close to the oracle (best weight) on RT-09 (7.2\% $\rightarrow$ 7.4\%).

\subsubsection{Initial cluster number}

\begin{table}[tb]
    \centering
    \caption{Speaker errors with varying initial cluster number}
    \label{tab:results-nclusters}
    \begin{tabular}{c|c|c}
        \hline
         Initial no.~clusters    & RT-07     & RT-09 \\
                        & (dev)     & (test) \\
         \hline
            16 (default)&   4.1     &   8.5 \\
            18          &   2.6     &   7.4 \\
            20          &   \bf 2.5     &   7.4 \\
            22          &   3.1     &   7.2 \\
            24          &   5.5     &   \bf 6.7 \\
            \hline  
            DOVER       &   \bf 2.1     &   \bf 6.5
    \end{tabular}
\end{table}

Table~\ref{tab:results-nclusters} shows the results from varying the number of initial clusters in the ICSI algorithm.
The TDOA stream weight was fixed at 0.75, with 5 Gaussians per initial cluster.
The number of clusters was increased in steps of two starting 
from the old default value of 16.

Again we find a non-smooth dependency of the dev error rate on the parameter value, and poor correlation between dev and eval results.
In fact, the worst choice for the dev set (24 initial clusters) gives the best result on the eval set.
The DOVER combination improves on the best single parameter choice for both RT-07 (2.5\% $\rightarrow$ 2.1\%) and RT-09 (6.7\% $\rightarrow$ 6.5\%).

\subsubsection{Initial Gaussian number}

\begin{table}[tb]
    \centering
    \caption{Speaker errors with varying initial number of Gaussians}
    \label{tab:results-ngaussians}
    \begin{tabular}{c|c|c}
        \hline
         Initial no.~of Gaussians  & RT-07     & RT-09 \\
                        & (dev)     & (test) \\
         \hline
            3           &   \bf 3.0     &   \bf 7.7 \\
            4           &   \bf 3.0     &   \bf 7.7 \\
            5 (default) &   4.1         &   8.5 \\
            10          &   5.8        &   7.3 \\
            \hline  
            DOVER (3,4,5) &   \bf 3.0     &   \bf 7.7 
    \end{tabular}
\end{table}

Table~\ref{tab:results-ngaussians} shows the results from
varying the number of Gaussians allocated to initial cluster models for the MFCC features (the TDOA features always receive one initial Gaussian per cluster, consistent with a unimodal distribution).
The TDOA weight was fixed at 0.75 and the initial cluster number was 16.

Here, for once, the eval set results seem to track the dev set nicely, and a close to optimal result can be achieved by picking a hyperparameter value on the dev set (3 or 4 Gaussians).
The only countervailing result is that an even better eval result could have been achieved by using 10 Gaussians, a choice that
gives very poor results on the dev set.
DOVER combination of the three choices with reasonable devset results (3, 4, 5) gives a result that is consistent with a majority vote among the three inputs.

\subsection{Randomization}

Following the randomization algorithm outlined in Section~\ref{sec:randomization}, a pseudo-random number between 0 and 1 is generated at each iteration of the clustering algorithm.
If the second-best merge pair still produces a positive change in the likelihood, {\em and}
the random value is less than $p = 0.3$, the second-best merge is chosen over the first-best merge.
For each meeting the diarization was run five times with changing random seed values to generate diverse outputs, which were then combined with DOVER.
(An odd number of trials was chosen to make ties less likely at the DOVER voting stage.)

\begin{table}[tb]
    \centering
\caption{Results with standard best-first clustering and randomized second-best clustering under different random seeds.  The final row represents the DOVER combination of the randomized trials.}
    \label{tab:results-random}
    \begin{tabular}{l|c|c|c}
        \hline
        Method      & Seed  & RT-07 & RT-09     \\
        \hline
        Best first  &       & \bf 4.1   & \bf 8.5       \\
        \hline
        Randomized  &  1    & 2.8   & 7.5       \\
                    &  2    & 2.4   & 8.1       \\
                    &  3    & 3.7   & 8.3       \\
                    &  4    & 3.6   & 8.7       \\
                    &  5    & 5.4   & 8.5       \\
        \hline 
        DOVER       &       & \bf 3.3   & \bf 8.1       \\
    \end{tabular}
\end{table}

Results for all experiments are shown in Table~\ref{tab:results-random}.
An immediate observation is that results vary quite a bit with randomized second-best merge picking, with both higher and lower error rates than obtained with the standard best-first clustering.
The range from best to worst is 73\% of the baseline for RT-07,
and 14\% for RT-09.

Of course there is no predicting which random seed gives the best results (and there is no possible correlation between development and test performance for a given seed).
However, DOVER-voting among all five outputs reduces speaker error by 20\% relative on RT-07, and by 5\% relative on RT-09, relative to the best-first baseline.

\section{Conclusions and Future Work}
	\label{sec:concl}

We have presented an novel approach for making diarization outputs more robust to choice of hyperparameters and greedy optimization,
a problem that is pronounced for algorithms based on
best-first agglomerative clustering.
The DOVER algorithm is used to vote among the multiple diarization outputs generated from the same input.
Random choice of second-best merges is one technique to mitigate the greediness of the clustering decisions, and in combination with DOVER averaging, yields gains over the best-first clustering algorithm.

Among the hyperparameters investigated, the relative weighting of features streams and the initial number of clusters are shown to be very difficult to optimize, both because of non-smooth results and poor generalization across test sets, as the hyperparameters are varied. 
In this situation, DOVER improves robustness, and often the overall best achievable result, by averaging diarization outputs over multiple hyperparameter settings.

Future exploration of the ideas presented here could include
varying hyperparameters along multiple dimensions jointly and
other forms of randomization within the logic of the clustering algorithm.
Also, we plan to adapt the concepts of hyperparameter variation and randomization to more recent diarization approaches, such as those based on
variational Bayes \cite{DiezEtAl:interspeech2019},
spectral clustering \cite{WangEtAl:icassp2018} and neural models \cite{YinEtAl:interspeech2018,FujitaEtAl:interspeech2019}.

\section{Acknowledgments}
Thanks to colleagues in the Microsoft diarization v-team for
fruitful discussions, to Xavi Anguera for information on the use of TDOAs in diarization, and to ICSI for making the data from the NIST RT evaluations available.

\bibliographystyle{ieee-shortnames}
\bibliography{refs}

\end{document}